\documentclass[letterpaper, 10 pt, conference]{ieeeconf}  

\IEEEoverridecommandlockouts                              

\usepackage{times}
\usepackage{epsfig}
\usepackage{graphicx}
\usepackage{amsmath}
\usepackage{hyperref}
\usepackage{algorithm}
\usepackage[noend]{algpseudocode}

\usepackage{subcaption}
\DeclareCaptionSubType*[Alph]{table}
\DeclareCaptionLabelFormat{mystyle}{Table~\bothIfFirst{#1}{ ̃}#2}
\usepackage{multirow}
\usepackage{array}
\newcolumntype{P}[1]{>{\centering\arraybackslash}p{#1}}

\usepackage{xcolor}
\usepackage{comment}

\title{\LARGE \bf
Interactive Continual Learning Architecture for Long-Term Personalization of Home Service Robots
}

\author{Ali~Ayub$^{1*}$, Chrystopher L.\ Nehaniv$^{1}$,  and Kerstin~Dautenhahn$^{1}$
\thanks{This research was undertaken, in part, thanks to the Canada 150 Research Chairs Program.}
\thanks{$^{1}$University of Waterloo, Waterloo, ON N2L 3G1, Canada}
\thanks{{\tt\small \{$^*$a9ayub, cnehaniv, kdautenh\}@uwaterloo.ca}}
}

\begin{document}

\maketitle
\thispagestyle{empty}
\pagestyle{empty}

\begin{abstract}
\label{sec:Abstract}
For robots 
to perform assistive tasks in unstructured home environments, they must 
learn and reason on the semantic knowledge of the environments. Despite a resurgence in the development of semantic reasoning architectures, these methods assume that all the training data is available a priori. However, each user's environment is unique and can continue to change over time, which makes these methods unsuitable for personalized home service robots. Although research in continual learning develops methods that can learn and adapt over time, most of these methods are tested in the narrow context of object classification on static image datasets. In this paper, we combine ideas from continual learning, semantic reasoning, and interactive machine learning literature and develop a novel interactive continual learning architecture for continual learning of semantic knowledge in a home environment through human-robot interaction. The architecture builds on core cognitive principles of learning and memory for efficient and real-time learning of new knowledge from humans. We integrate our architecture with a physical mobile manipulator robot and perform extensive 
system evaluations in a 
laboratory environment over two months. 
Our results demonstrate the effectiveness of our architecture to allow a physical robot to continually adapt to the changes in the environment from limited data provided by the users (experimenters), and use the learned knowledge to perform object fetching tasks. 
\end{abstract}

\section{Introduction}
\label{sec:introduction}
\noindent Significant advances in key robotics technologies, such as object recognition and object detection~\cite{Russakovsky15,Girshick_2015_ICCV}, robot navigation and localization~\cite{fulda2017harvesting,wang2018automatic}, robot scene understanding~\cite{Ayub_2020_BMVC}, and semantic reasoning~\cite{liu2023survey}, have enabled home service robots to perform daily tasks in household environments. To be able to plan and perform tasks in an unstructured home environment, the first step is for the robot to understand the environment i.e. objects (e.g. cup, bottle), contexts (e.g. kitchen), and semantic relations between objects and contexts (e.g. cup is in a cupboard in the kitchen). 
However, each user's environment can be unique, and 
it can continue to change over time, e.g. a user might replace old objects with new ones, or change objects' locations. In such dynamic environments, developing general-purpose robots might not be the best solution, instead, a personalized approach~\cite{Saunders16} to developing home service robots might be more appropriate, where these robots can interact with their users to learn about their unique environments. A combination of interactive machine learning~\cite{thomaz09} and continual learning (CL)~\cite{Rebuffi_2017_CVPR} 
is a promising solution to this problem, allowing users to interactively teach their personal home service robots about their environments through repeated interactions over time.

CL has been the subject of growing interest in recent years, to develop machine learning (ML) models that can learn and adapt over extended periods of time~\cite{Rebuffi_2017_CVPR,Ayub_2020_CVPR_Workshops,hayes2021replay}. This research has found valuable applications in the field of robotics, where it enables robots to consistently acquire knowledge about diverse objects, leveraging limited data inputs often provided by their users~\cite{Ayub_IROS_20,ayub2023roman}. Despite significant advancements, most CL methods are developed for object recognition, which is crucial but insufficient to 
perform home service tasks. Although a few recent CL methods were developed for knowledge graph embeddings for semantic object reasoning, these methods were only developed for symbolic datasets without any application to physical robots where new knowledge might not be provided in constrained setups used for evaluating these methods~\cite{daruna2021continual,cui2023lifelong}. Further, common across most CL methods for robotics is that these methods are not developed to learn from real-time interactions with their users. 
Although interactive ML methods have been developed that can allow robots to interact and learn object categories~\cite{thomaz09}
, these methods are only tested in single interaction scenarios and not over long-term interactions for CL. 

For understanding unstructured environments beyond object recognition, semantic reasoning architectures~\cite{liu2023survey,wang2020home,daruna2021towards,liu2023learning} have been developed for home service robots. Albeit promising advances in 
this field, these methods use static knowledge sources that are trained on a large corpus of symbolic datasets without much personalization or any interaction with their users~\cite{wang2020home,daruna2021towards,liu2023learning,liu2022service}. To the best of our knowledge, 
we know of no other work on integrating modern CL methods with semantic reasoning architectures that can allow physical robots to learn continually over a long period of time through human-robot interaction (HRI).



In this paper, we combine ideas from CL, interactive ML, and semantic reasoning literature to develop a novel interactive continual learning (ICL) architecture for a personalized home service robot that can interact with, learn from, and assist its users in their dynamic home environments.
Our ICL architecture allows users to continually teach a home service robot the semantic knowledge of the environment i.e. objects and contexts in the home environment, and allows the robot to use the dynamic knowledge about the environment to perform object search and fetching tasks. 
Further, as users might provide limited data and expect quick results, our ICL architecture can learn in real time from only a few training examples of objects and contexts provided by the users. We perform extensive proof-of-concept evaluations of the ICL architecture on a Fetch mobile manipulator robot~\cite{Wise16} over two months to continually learn 20 common household objects and two home contexts. Our results demonstrate the ability of our architecture to quickly adapt to the environment with limited data without catastrophically forgetting any past knowledge. This paper makes the following contributions:

\begin{figure*}[t]
\centering
\includegraphics[width=0.7\linewidth]{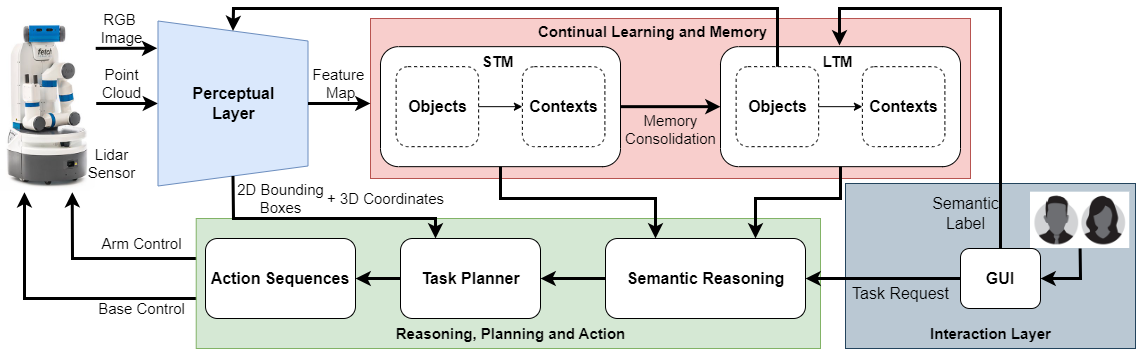}
\caption{\small Overview of the ICL architecture. It is composed of 1) perception, 2) interaction, 3) continual learning and memory, and 4) reasoning, planning, and action.}
\label{fig:icl_architecture_fig}
\end{figure*}

\begin{itemize}
    \item We 
    propose a novel ICL architecture that integrates CL, interactive ML, and semantic reasoning to address multiple real-world challenges that have only been partially addressed in prior research, i.e. it can continually learn both objects and contexts through HRI in real-time with limited data, without catastrophically forgetting past knowledge. 
    \item Unlike most prior CL research, we perform extensive CL experiments on a physical robot in a large indoor environment. Our results demonstrate the real-world applicability of our ICL architecture to adapt to its environment and perform object fetching tasks. 
\end{itemize}

\section{Related Work}
\label{sec:related_work}
\subsection{Continual Learning}
Continual learning (CL) models aim to achieve a dual objective: to continually adapt and learn new information over time while simultaneously preserving 
past knowledge~\cite{Rebuffi_2017_CVPR,hayes2021replay,Ayub_iclr20,shi19b}. In the literature, various categories of CL scenarios are defined based on factors such as shifts in input or output distributions and whether inputs and outputs share the same representation space~\cite{lomonaco17}. Among these categories, class-incremental learning (CIL) is the most studied problem, in which an ML model is trained using labeled training data of different classes in each increment and is then evaluated on all the classes it has learned up to that point~\cite{Rebuffi_2017_CVPR}. One of the primary challenges encountered by CIL models is \textit{catastrophic forgetting}, a phenomenon in which the CIL model forgets the previously learned classes when it is exposed to new classes in an increment~\cite{french19}. Numerous research avenues have been explored in the quest to mitigate catastrophic forgetting. These include replay-based techniques~\cite{Rebuffi_2017_CVPR,hayes2021replay,Wu_2019_CVPR}, where data from previously learned classes is stored and replayed during the learning of new classes, regularization techniques~\cite{kirkpatrick17,Li18} that constrain model parameters from changing drastically, and generative replay-based techniques~\cite{Ayub_iclr20,Ostapenko_2019_CVPR} which generate data for old classes using stored class statistics. Common across these CL techniques is the requirement of a large number of training examples per class, making them unsuitable for learning through HRI. 
In recent years, CL researchers developed models that can learn from only a few labeled training examples per class, a direction called few-shot class incremental learning (FSCIL)~\cite{tao_2020_ECCV}. All of the CIL and FSCIL techniques were designed for the narrow context of object classification only, and mostly tested on static datasets in systematically controlled setups and not on physical robots learning in real-time through HRI~\cite{tao_2020_ECCV,Ayub_2020_CVPR_Workshops,Ayub_IROS_20,Dehghan19,ayub2022few}.

\subsection{Semantic Reasoning Architectures} 
For real-world applications, robots need to reason on the semantic knowledge of objects (e.g. object categories, relations with other objects, contexts, etc.) to perform everyday tasks, such as object search and manipulation~\cite{wang2020home,daruna2021towards,liu2022service,scheutz17}, and navigation in dynamic and unknown environments~\cite{fulda2017harvesting}. Extensive research has been conducted in the past on semantic reasoning architectures to learn semantic knowledge of objects and environments, and reason on the 
mathematical relationships between different semantic properties of an object and between different objects~\cite{liu2023survey,wang2020home,daruna2021towards,liu2023learning}. Most of these architectures use static knowledge sources~\cite{liu2023survey} 
that are trained and tested on a large corpus of semantic reasoning datasets, which might not be personalized to unique user environments. A few semantic reasoning architectures have been tested on physical robots in constrained setups for object search tasks, however, they were not tested over the long term in dynamic environments.   

Although recently a few preliminary research directions have been considered for developing CL methods for graph embeddings~\cite{daruna2021continual,cui2023lifelong}, these methods simply apply CL methods for object classification on knowledge graphs. Further shortcomings of these methods are: 1) semantic CL methods continually learn knowledge graph embeddings only from symbolic data and do not learn object categories (from image data) simultaneously with the semantic knowledge embeddings, 2) these methods are computationally expensive and tested on static datasets in constrained CL setups without any interaction with the users. In this paper, we develop a novel ICL architecture to continually learn both objects and contexts in a dynamic environment over the long term. 

\section{Interactive Continual Learning Architecture}
\label{sec:methodology}
\noindent 
We build 
an interactive learning framework that includes some of the core 
abilities of a cognitive architecture, mainly perception, action, learning, adaptation, autonomy, memory, and reasoning. 
The complete interactive continual learning (ICL) architecture is depicted in Figure~\ref{fig:icl_architecture_fig}. 
All the components in ICL architecture are implemented as ROS 
nodes, 
communicating through topic subscriptions.

\subsection{Perception Layer}
\label{sec:perception_layer}
The perception layer manages the robot's perception capabilities to extract information from the visual input from the robot's RGB-D camera. Given that there can be many objects within a home environment, the first step is to individually detect each object in the robot's camera view. 
For a home service robot, 
an object detection algorithm must be chosen based on real-time constraints, as users would 
want the robot to learn and perform object fetch tasks efficiently. The most used and appropriate algorithm for real-time object detection is YOLO~\cite{Redmon_2016_CVPR}. We use a pre-trained YOLO for detecting objects in the RGB image taken from the robot's camera. The pre-trained YOLO can also provide object categories, however, these predicted categories are not useful as the objects in a user's home might be different from the pre-trained categories, and even for the same objects a user might want to label objects differently 
\cite{ayub2023roman}. Given that there will only be a few training examples provided by the users, following the common practice in few-shot learning~\cite{Chen19} and FSCIL~\cite{Ayub_2020_CVPR_Workshops} literature, the detected objects are further processed through a pre-trained CNN on ImageNet~\cite{Russakovsky15} for extracting latent features. The latent features can then be used for continual learning of objects. The point cloud data obtained from the depth sensor is combined with the detected objects in the RGB image to determine the 3D location of the object so that the manipulator can reach and pick the objects. A lidar sensor is also used for the robot's base to navigate in the environment while avoiding obstacles.    


\subsection{Interaction Layer}
\label{sec:interaction_layer}
A key aspect of the ICL architecture is interaction with the users to allow them to teach the robot new knowledge and changes in the environment, and also request the robot to perform object fetch actions. 
We developed a graphical user interface (GUI) that allows the user to teach the robot objects and contexts in the environment. For teaching an object, the user can place the object in front of the robot's camera, and type the label of the object in a textbox. Similarly, for teaching a context, say a kitchen, the user can manually drive the robot using a joystick to the context location and then type the name of the context and the location number in a text box. The robot captures the images in its camera view which are then processed by the perception layer (Section~\ref{sec:perception_layer}). The user can also initiate an object fetch task by typing the name of the object to be fetched in a textbox. Examples of how the GUI is used for teaching and testing are shown in the supplementary video (\url{https://youtu.be/9y7zf4JUQGU}).

Similar to other layers in the architecture, the interactive technology (GUI) in this layer can be replaced. For example, a natural language interaction system, as used by many general-purpose cognitive architectures (e.g. SOAR~\cite{laird1987soar}) can be added. However, to avoid any speech-to-text and natural language understanding errors during learning, we opted to use a GUI. 

\subsection{Continual Learning and Memory}
At any instant in time, the user can start a teaching interaction with the robot to either teach a new object, show more viewpoints of a previously shown object, or teach a part of the context (e.g. an area of a kitchen). The underlying principles for learning object categories and contexts are similar. Therefore, we will first explain the continual learning and memory development procedure for object recognition and then describe some differences in learning contexts. 

\subsubsection{Continual Learning}
\label{sec:continual_learning}
At any time instant $t$, the user can teach the robot objects belonging to $n^t$ different categories. Let's consider that the robot has learned $N^{t-1}$ number of object categories before. In this unconstrained setup, new objects can belong to any of the previously learned categories, which is different from the constrained class-incremental learning~\cite{Rebuffi_2017_CVPR} setting where data for an object category is only available once. To teach the robot a new object belonging to category $i$, the user shows $m_i$ number of object views to the robot. After processing the images through the perception layer, the architecture receives $m_i$ number of CNN feature maps for the object. For CL, we use a network architecture similar to SUSTAIN~\cite{love04} that clusters similar stimuli together. This network is inspired by the theories of pattern separation and memory integration in computational neuroscience~\cite{Mack17}. For each object belonging to category $i$ in increment $t$, the network begins with a simple solution and initiates a single cluster represented by the cluster centroid $c_1^i$ computed as a mean of the feature maps in the cluster. Let's assume that there are $k_i$ number of clusters already in memory for object $i$. For each new $j$th feature map $x_j$ for object $i$ in increment $t$, L1 distance is computed between $x_j$ and each of the $k_i$ cluster centroids for object $i$ over all the feature dimensions. Based on the distance from each cluster, a cluster activation is determined:

\begin{equation}
    H_l^{act} = e^{-d_l/w_l}
\end{equation}

\noindent Where $d_l$ is the L1 distance between the $l$th centroid of object category $i$ and $x_j$, and $w_l$ is the weight 
attached to cluster $l$. In our experiments, we set $w_l$ to be the total number of feature maps in the cluster.  Clusters compete with each other to respond to the input pattern (the feature map $x_j$), and the winning cluster is selected. For the winning cluster, say $l_w$th cluster, the output is calculated as,

\begin{equation}
    H_{l_w}^{out} = \frac{(H_{l_w}^{act})^\beta}{\Sigma_{s=1}^{k_{i}}(H_s^{act})^\beta} H_{l_w}^{act}
\end{equation}

\noindent Where $\beta$ is an inhibition parameter for the regulation of cluster competition (for our experiments, we set it to 1). Note that based on the above equation, if many clusters are strongly activated, the output of the winning cluster is less, because of inhibition by competing clusters. In this network, only the winning cluster is selected and the output for all the other clusters is set to zero. As mentioned earlier, this network starts with a simple solution including one cluster only but recruits more clusters if needed. The cluster recruitment process is self-supervised. If the winning cluster's output ($H_{l_w}^{out}$) is below a certain threshold ($\tau$) a new cluster is recruited, 
and the centroid of the cluster is set to the input feature map. Otherwise, if the output is higher than the threshold, the cluster centroid is updated:

\begin{equation}
    c_{l_w}^i = \frac{w_{l_w}c_{l_w}^i + x_j^i}{w_{l_w}+1}
\end{equation}

\noindent The weight $w_{l_w}$ for the winning cluster is also incremented by 1. Since only the winning cluster is updated for each feature map, this effectively prevents cluster interference, thereby mitigating catastrophic forgetting. Further, as the network requires only a single pass of training data, the learning process is efficient and can be executed in real time, which is a crucial component for learning through HRI. 

\subsubsection{Long-Term Memory}
\label{sec:ltm}
The network learned through CL (Section~\ref{sec:continual_learning}) is stored in the long-term memory (LTM) with all the clusters and the associated weights. Although the goal of CL is to avoid catastrophic forgetting, memory fading does happen in humans, especially for events that are not encountered regularly~\cite{collins2020improving,walsh2018evaluating}.  For a home environment, it is possible that some objects might simply be replaced over time and the previously used objects are no longer in the home. In such cases, it would be useless to store this knowledge in memory. As the robot would not encounter any replaced objects in its environment, we incorporate a memory-fading process in the LTM. We use the predictive performance equation (PPE) model~\cite{walsh2018evaluating,collins2020improving} of learning and retention. Based on this model, we represent each weight $w_i$ for cluster $i$ in our network as:

\begin{equation}
    w_i = w_i^\eta \times T^{-\alpha}
\end{equation}

\noindent Where, $\eta$ is the learning rate (held at 1 in~\cite{collins2020improving}), $\alpha$ is the decay rate, and $T$ is called the \textit{Model Time} in the PPE model, which is a weighted sum of the time since each of the $n$ previous events $T = \sum_{j=1}^{n} h_jt_j$, where $t_j$ is the time since the $j$th event. In our architecture, $t_j$ represents the time since cluster $i$ was activated the $j$th time. $h_j$ is the weight assigned to each event which decreases with time:

\begin{equation}
    h_j = \frac{t_j^{-u}}{\Sigma_{k=1}^{n}t_k^{-u}}
\end{equation}

\noindent Where $u$ controls the steepness of weighting, with higher values resulting in larger weights for more recent events. The weights $h_j$ are normalized and sum to 1. Through this process, the weights of clusters that are not refreshed (details below in Section~\ref{sec:stm}) are faded over time. We choose the decay rate $\alpha$ to be small in LTM to allow for slow fading.

\subsubsection{Short-Term Memory}
\label{sec:stm}
Inspired by the dual-memory system in the mammalian brain~\cite{mcclelland95,kitamura2017engrams}, our ICL architecture also contains a short-term memory (STM). Although the knowledge learned from the users is stored in LTM, STM stores the daily experiences of the robot encountered through its sensors and processed through the perception layer. STM stores the feature maps of encountered objects (and contexts) while performing tasks in the environment. Note that the clustering network is not used for STM because STM stores specific experiences, in accordance with the dual memory theory~\cite{mcclelland95}, whereas the LTM stores abstract knowledge represented as clusters. A weight parameter for each specific object instance (and context) is also stored to keep track of how many times the object was encountered. If the weight parameter exceeds a certain threshold $\gamma$ (a hyperparameter), the corresponding feature map is consolidated into LTM through the learning process described in Section~\ref{sec:continual_learning}. The memory consolidation process refreshes LTM and prevents relevant object and context memories from fading in LTM. The memory fading process described in Section~\ref{sec:ltm} is also applied to the weights of the feature maps stored in STM. The decay rate $\alpha$ for STM is chosen to be much higher as STM fades quickly compared to LTM. The data stored in STM can be used for assistive tasks beyond object fetching, for example, helping users find a misplaced object in the home~\cite{shah_hri_23} or if they forgot to perform an important task, such as taking their medicine. We will expand our architecture for these applications in the future. 

The above-mentioned process (Sections~\ref{sec:continual_learning}~\ref{sec:ltm}~\ref{sec:stm}) of learning and memory is described for object category learning. For context learning, the user would show a specific view of the context that can have multiple objects. As contexts can be described by the objects present in the scene, the architecture first generates the feature maps of all the objects in the image and then uses the object category network stored in LTM to find object categories in the context image. Following the principle of conceptual spaces~\cite{gardenfors2004conceptual}, we generate a feature map where the dimensions of the feature map represent the object categories present in the context image, and the location of the context. After the feature map is generated, a context cluster network is trained using the CL process described in Section~\ref{sec:continual_learning}. The LTM and STM for the context data also behave in the same way as for objects. Note that context recognition is dependent on object recognition, which means that the user would have to teach the objects first before they are shown in a context. However, a self-supervised learning method can be used to learn a new object in a context if it is not taught by the user. 

In summary, the learning and memory mechanisms within the ICL architecture are inspired by 
cognitive science principles, and they enable efficient real-time learning of objects and contexts with limited data while addressing the challenge of catastrophic forgetting. Nevertheless, similar to the other layers, it's possible to substitute the CL method and memory components with alternative techniques in the literature.

\subsection{Reasoning, Planning, and Action}
The ICL architecture allows a mobile manipulator robot to perform object fetch tasks in a home environment. For this task, the architecture receives the object to be fetched from the user through the interaction layer (Section~\ref{sec:interaction_layer}). A masked conceptual space feature map is generated using the requested object category and passed through the 
context learning network in LTM which predicts the most probable context location for the object. The context location is then combined with a metric map (SLAM map~\cite{csorba1997simultaneous}) of the environment, and a navigation planner in ROS is used to plan a collision-free path to the context location. The robot then moves to the context location. As there can be multiple objects present at a context location, the architecture localizes the objects using the perception layer (Section~\ref{sec:perception_layer}) and gets the predicted object labels from the object learning network in LTM. 
If the object requested by the user is in the list of detected objects at the context location, the architecture plans an arm trajectory to move to the 3D coordinates of the object (from the perception layer). The robot's arm then moves to the object's location and picks it up. The navigation planner then plans a path back to the user's location and the robot's base moves to that location. Finally, the robot places the object on the table using the arm trajectory planner. 

\section{Experiments}
\label{sec:experiments}
\noindent In this section, we describe the experimental setup of the ICL architecture on a physical robot and the object fetch experiment to evaluate ICL's performance. The researchers assume the role of users throughout this section.

\subsection{Experimental Setup}
\begin{figure}[t]
    \centering
    \begin{subfigure}{.5\linewidth}
        \includegraphics[width=\linewidth]{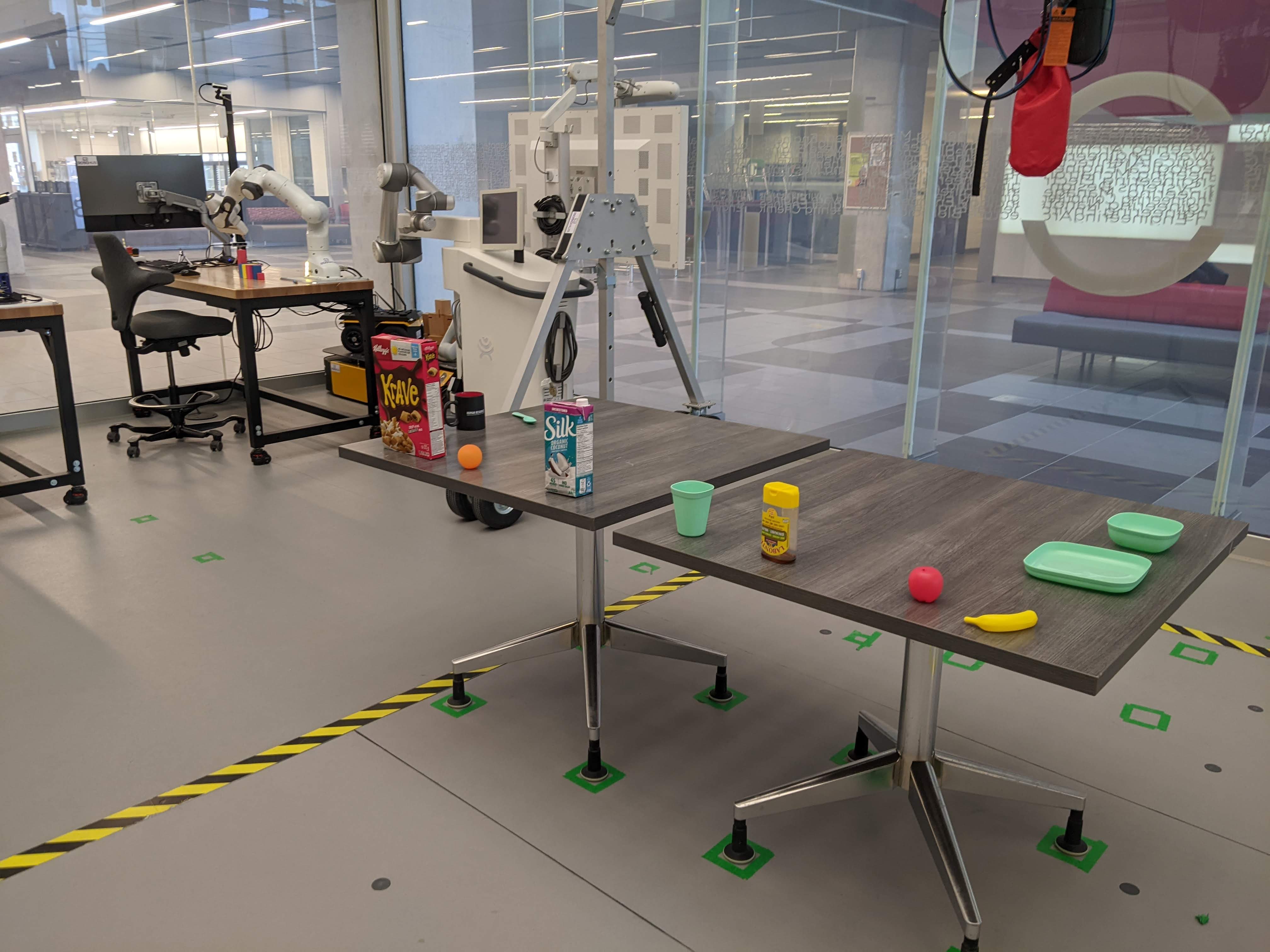}
        \caption{}
        \label{fig:kitchen:a}
    \end{subfigure}%
    \begin{subfigure}{.5\linewidth}
        \includegraphics[width=\linewidth]{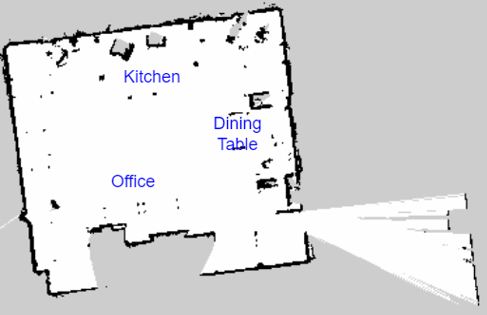}
        \caption{}
        \label{fig:kitchen:b}
    \end{subfigure}
    \caption{(a) An example of the kitchen context in our experimental setup. Note the variation in object size, the closeness of objects, and distractions, such as robots and transparent walls behind. See the supplementary video (https://youtu.be/9y7zf4JUQGU) for examples of other contexts. (b) SLAM map of the indoor environment.}
    \vspace{-1.6em}
    \label{fig:kitchen}
\end{figure}

We evaluate our architecture on a Fetch mobile manipulator~\cite{Wise16} equipped with a 7 DOF arm, a mobile base, a lidar, and an RGB-D camera sensor. For faster object detection and CL, an external workstation equipped with an Nvidia RTX 2060 GPU is used. We set up three mock home contexts (kitchen, home\_office, and dining\_area) in an indoor laboratory environment. 
We argue that our lab setup is comparable to mock home setups in previous work~\cite{jiang2019open,wang2020home,liu2022service,daruna2021towards,liu2023learning} due to the similar use of table-top environments for object placement. Further, prior works used only a few objects ($<$10) in physical robot experiments with objects placed individually without many other objects close to each other to avoid detection failures. Unlike these works, in our experimental setup, we use 20 different objects of various sizes and shapes that are commonly found in home environments, and we place them close to each other in the contexts (see Figure~\ref{fig:kitchen:a} for an example). We map the indoor environment using the lidar sensor and store the SLAM map in the robot's onboard memory (Figure~\ref{fig:kitchen:b}). 

The hyperparameters for the CL method and LTM and STM are set as follows: the threshold $\tau$ in the CL network for recruiting a new cluster is set to 0.0003 for the object recognition network, and it is set to 0.13 for context recognition network. These thresholds were chosen using cross-validation on object and context data in the first increment. The decay rate $\alpha$ is set to 0.2 for LTM, and 15 for STM. These decay rates were chosen such that it would take almost one month for the weights in LTM to decay to 50\% their original value ensuring slow fading, and 2 days for the weights in STM to decay to $\sim$0.003\% leading to complete memory fading. As we performed a few runs of the experiment each day, we measured the time since the activation of a cluster in days. The variable $u$ in equation (5) was set to 0.6 as in~\cite{walsh2018evaluating}. Finally, we use a ResNet18~\cite{He_2016_CVPR} as the feature extractor in the perceptual layer, as it is used commonly for other CL methods~\cite{tao_2020_ECCV,Zhang_2021_CVPR,Ayub_IROS_20}. 

\subsection{Continual Learning Experiment}
We performed a long-term CL experiment to teach the robot objects and contexts continually and test it on object fetch tasks. The object fetch tasks can help us evaluate both the object and context learning pipelines given that both are needed to successfully find the object in the environment.  In this experiment, we taught the object and context data to the robot over 16 increments. Each increment was usually a day apart, with some days consisting of two increments. For two increment days, we updated the event time in memory fading twice. For robustness, we performed this experiment 5 times randomizing the order of objects and contexts taught to the robot. Out of the 20 objects, 7 were assigned to the office and 13 to the kitchen. In each increment, the robot was either taught some objects only, some context updates only, or both objects and contexts. For the first 10 increments, after each increment, we tested the robot 5 times to fetch some of the objects it had learned so far. By 10 increments all 20 objects were taught to the robot, therefore, for the last 6 increments, only contextual updates were taught to the robot where some objects were randomly moved within contexts or completely removed. In the last 6 increments, we tested the robot 7 times each to fetch objects it had learned so far. 

We report the task execution accuracy representing if the robot was able to fetch the correct object in each increment. A successful task execution depends on other components of ICL, therefore we report the accuracy of each of the components. 1) object classification accuracy (\%) measures the percentage of correctly classified objects when the robot moves to context locations, 2) context classification accuracy (\%) measures the correct prediction of context location for the requested object, 3) manipulation error measures the number of times the robot was unable to physically pick, place, or move objects, 4) perception error measures the object detection failures in the perception layer, 5)  navigation error measures the number of times the robot's base failed to move to the desired location, and 6) execution time measures the time required by the robot to correctly fetch an object. Given the continual nature of our experiment, and the time required to teach and test the robot, it took approximately 2 months to complete 5 experimental runs (16$\times$5 increments). 

Table~\ref{tab:accuracy_results} shows results for the first 10 increments averaged over 5 runs of the experiment. The task execution accuracy starts at $\sim$80\% and decreases to $\sim$56\% ($\sim$24\% decrease) after learning 20 objects and two contexts in 10 increments, therefore we notice some forgetting of past knowledge but no catastrophic forgetting. Diving deeper, we notice that the main source of task accuracy decrease is the object recognition network, where the accuracy starts at $\sim$100\% but decreases to $\sim$73\% ($\sim$27\% decrease). The lower object recognition accuracy, however, could be because the network has more classes to classify, and the limited amount of data provided per object. On average, only \textbf{5-10 images per object} were shown to the robot for training in the experiment. We also trained the ICL architecture on all the object and context data at once (called joint training (JT)) and performed the object fetch task 15 times. The difference in task execution accuracy of ICL in the 10th increment compared to JT was only 6.1\%, which indicates that ICL mitigates catastrophic forgetting effectively. The context recognition network maintained almost perfect accuracy for all increments without any forgetting. The only errors in the context recognition module were because of one misclassified object during the learning phase of the contexts. The object \textit{fork} was predicted as \textit{spoon} by the object recognition network during kitchen context learning. The other sources of errors in task execution were the manipulation and perception errors, both of which were consistently $\sim$10-20\% for all increments as these errors did not depend on object or context learning networks. Finally, the navigation stack performed quite reliably, albeit with a few failures in each experimental run.  

\begin{table}[t]
    \centering
    \begin{tabular}{c | c | c | c | c}
         \textbf{} & \textbf{Objects} & \textbf{Contexts} & \textbf{Tasks} & \textbf{Execution Time} \\ \hline
        1 & 100.0 $\pm$ 0 & 100 $\pm$ 0 & 80.0 $\pm$ 0.1 & 96.6 $\pm$ 3.6 s \\
        2 & 93.1 $\pm$ 5.5 & 96.0 $\pm$ 0.1 & 72.0 $\pm$ 0.1 & 96.0 $\pm$ 4.6 s\\ 
        3 & 91.8 $\pm$ 7.3 & 96.0 $\pm$ 0.1 & 72.0 $\pm$ 0.1 & 93.0 $\pm$ 3.2 s\\
        4 & 88.7 $\pm$ 5.0 & 100 $\pm$ 0 & 70.0 $\pm$ 0.1 & 98.2 $\pm$ 2.3 s\\
        5 & 85.6 $\pm$ 4.8 & 92.9 $\pm$ 0.1 & 66.0 $\pm$ 0.2 & 99.0 $\pm$ 3.5 s\\
        6 & 83.7 $\pm$ 4.3 & 100 $\pm$ 0 & 64.0 $\pm$ 0.2 & 97.1 $\pm$ 3.4 s\\
        7 & 80.1 $\pm$ 3.7 & 96.0 $\pm$ 0.1 & 62.0 $\pm$ 0.2 & 98.4 $\pm$ 2.6 s\\
        8 & 78.2 $\pm$ 3.4 & 96.0 $\pm$ 0.1 & 62.0 $\pm$ 0.2 & 100.6 $\pm$ 3.9 s\\
        9 & 75.2 $\pm$ 2.3 & 100 $\pm$ 0 & 58.0 $\pm$ 0.3 & 99.4 $\pm$ 2.4 s\\
        10 & 73.3 $\pm$ 0.6 & 100 $\pm$ 0 & 56.0 $\pm$ 0.3 & 97.5 $\pm$ 3.1 s \\
        \hline
        \textbf{JT} & \textbf{+3.7} & \textbf{+0} & \textbf{+6.1} & \textbf{+0}\\
    \end{tabular}
    \caption{Incremental task execution, object, and context recognition accuracy (\%), and task execution time (seconds) over first 10 increments averaged over 5 runs. JT represents the ICL architecture trained on all the data at once without any continual learning. The row for JT shows the accuracy and time differences compared to the 10th increment.}
    \vspace{-1.6em}
    \label{tab:accuracy_results}
\end{table}

For the \underline{last 6 increments}, no new objects were taught but the contexts were updated by moving or removing objects. In these runs, both the object classification and task execution \textbf{accuracies remained similar} to the accuracies in the 10th increment. The reason is that our context recognition module had almost zero forgetting and quick adaptation ability, which ensured similar performance for task execution given that all other sources of errors remained consistent. The reason for consistent context recognition accuracy was that as contexts were updated, older memories about those contexts continued to fade which resulted in updated context clusters winning during cluster competition in the context recognition network. In terms of the \underline{execution time}, on average it took \textbf{97 seconds} to fetch an object and place it on the dining table. The reason for the short execution time was that the robot went to the correct object locations without wandering around in the environment. For the last 6 increments, we also tested the architecture without any memory decay, which resulted in higher task execution times ($\sim$130 seconds, \textbf{33 seconds} higher) because the robot went to older locations of objects first which wasted some time. Finally, the \underline{training and testing time} for the CL networks were $\sim$1 second and 0.004 seconds, respectively, ensuring real-time learning and prediction capabilities of the architecture. Examples of teaching and object-fetching tasks are shown in the supplementary video (https://youtu.be/9y7zf4JUQGU).

\section{Conclusions}
\label{sec:conclusion}
\noindent In this paper we tackle the challenging problem of simultaneous continual learning of object and context knowledge in a dynamic home environment. We developed a novel ICL architecture that builds on the core cognitive principles of learning and memory 
and allows users to directly teach objects and contexts in their unique home environment. Our results on a mobile manipulator robot demonstrate the ability of ICL to continually adapt to the changes in the environment through limited data provided by the user. We hope that this work can serve as a first step toward developing personalized home service robots that can interact with, learn from, and perform assistive tasks for their users over the long term. 

\section{Limitations and Future Work}
\label{sec:limitations}
Given the long-term nature of the experiments with a physical robot, we did not perform an extensive hyperparameter analysis and compared our proposed approach with JT only. In the future, we hope to perform more experiments to analyze the tradeoff of hyperparameter values and use other SOTA CL models. In this paper, we only considered the supervised learning ability of ICL, however, in the future, we will expand it to perform self-supervised CL. Additionally, the experimental setup did not have overlapping objects in different contexts. In the future, we will analyze how ICL performs when there are overlapping objects in different contexts. Finally, we will conduct long-term HRI studies with human users in home environments to understand the personalization capabilities of ICL in the real world.






{\small
\bibliographystyle{IEEEtran.bst}
\bibliography{main}
}

\end{document}